\documentclass{article}

\usepackage{arxiv}

\usepackage[utf8]{inputenc} 
\usepackage[T1]{fontenc}    
\usepackage{hyperref}       
\usepackage{url}            
\usepackage{booktabs}       
\usepackage{amsfonts}       
\usepackage{nicefrac}       
\usepackage{microtype}      
\usepackage{lipsum}		
\usepackage{graphicx}
\usepackage{natbib}
\usepackage{doi}

\usepackage{times}
\usepackage{soul}
\usepackage{url}
\usepackage[small]{caption}
\usepackage{graphicx}
\usepackage{amsmath}
\usepackage{booktabs}
\usepackage{algorithm}
\usepackage{algorithmic}
\usepackage{amssymb} 
\usepackage{caption} 
\usepackage{booktabs}
\usepackage{subfigure}
\usepackage{multicol,lipsum}
\usepackage{multirow}
\usepackage{xcolor}
\usepackage{amsthm}
\DeclareMathOperator*{\argmax}{arg\,max}

\title{Unsupervised Path Representation Learning with Curriculum Negative Sampling}

\author{ Sean Bin Yang  \\
    Department of Computer Science \\
	Aalborg University, Denmark \\
	\texttt{seany@.cs.aau.dk} \\
	\And
	Chenjuan Guo \\
	Department of Computer Science \\
	Aalborg University, Denmark \\
	\texttt{cguo@.cs.aau.dk} \\
	\AND
	Jilin Hu\thanks{Corresponding Author: Jilin Hu} \\
	Department of Computer Science \\
	Aalborg University, Denmark \\
	\texttt{hujilin@.cs.aau.dk} \\
	\And
	Jian Tang \\
	Mila-Quebec AI Institute\\
    HEC Montreal, Canada\\
    CIFAR AI Research Chair \\
	\texttt{jian.tang@hec.ca} \\
	\And
	Bin Yang \\
	Department of Computer Science \\
	Aalborg University, Denmark \\
	\texttt{byang@.cs.aau.dk} \\
}

\renewcommand{\shorttitle}{\textit{arXiv} Template}

\hypersetup{
pdftitle={Unsupervised Path Representation Learning with Curriculum Negative Sampling},
pdfsubject={q-bio.NC, q-bio.QM},
pdfauthor={Sean Bin Yang, Chenjuan Guo, Jilin Hu, Jian Tang, Bin Yang},
pdfkeywords={First keyword, Second keyword, More},
}

\begin{document}
\maketitle

\begin{abstract}
Path representations are critical in a variety of transportation applications, such as estimating path ranking in path recommendation systems and estimating path travel time in navigation systems.
Existing studies often learn task-specific path representations in a supervised manner, which require a large amount of labeled training data and generalize poorly to other tasks. 
We propose an unsupervised learning framework
Path InfoMax (\emph{PIM}) to learn generic path representations
that work for different downstream tasks. 
We first propose a curriculum negative sampling method, for each input path, to generate a small amount of negative paths, by following the principles of curriculum learning. 
Next, \emph{PIM} employs mutual information maximization
to learn path representations from both a global and a local view. 
In the global view, \emph{PIM} distinguishes the representations of the input paths from those of the negative paths. In the local view, \emph{PIM} distinguishes 
the input path representations from the representations of the nodes that appear only in 
the negative paths. This enables the learned path representations encode both
global and local information at different scales. 
Extensive experiments on two downstream tasks, ranking score estimation and travel time estimation, using two road network datasets suggest that \emph{PIM} significantly outperforms other unsupervised methods and is also able to be used as a pre-training method to enhance supervised path representation learning. 
\end{abstract}

\keywords{Path Representation Learning \and Unsupervised Learning \and Mutual Information Maximization}

\section{Introduction}
Path representations are crucial for various transportation applications, e.g., travel cost estimation~\cite{DBLP:conf/icde/Hu0GJX20,DBLP:journals/pvldb/PedersenYJ20}, 
routing~\cite{DBLP:journals/vldb/GuoYHJC20,DBLP:journals/vldb/PedersenYJ20}, 
path recommendation~\cite{DBLP:conf/icde/Yang020,DBLP:conf/icde/Guo0HJ18}, 
and traffic analysis~\cite{DBLP:conf/icde/HuG0J19,razvanicde2021}. 
%
Path representation learning~(\emph{PRL}) aims to obtain distinguishable path representations for different paths in a transportation network 
and hence facilitating various downstream applications.  
Existing studies on \emph{PRL} 
often learn path representations in a supervised manner, which has two limitations. First, they require a large amount of labelled training data. 
Second, the learned path representations are task-specific, e.g., working well for the task with labels, but generalize poorly to other tasks. 
The two limitations restrict supervised path representation learning from broader usage, thus calling for unsupervised path representation learning. 

Although unsupervised graph representation learning methods exist, they are not designed to capture representations of paths. 
Node representation learning~\cite{DBLP:conf/www/TangQWZYM15,DBLP:conf/kdd/GroverL16} learns representations for individual nodes in a graph but does not consider paths, i.e., sequences of nodes. Simply aggregating the node representations of the nodes in a path fails to capture the sequential information in paths. 
Whole graph representation learning~\cite{DBLP:conf/iclr/SunHV020} learns representations for different graphs, while path representation learning considers different paths from the same graph. 
In addition, unsupervised graph representation learning often utilize random negative sampling to enable training, which is ineffective for path representation learning. 

We propose an unsupervised path representation learning framework Path InfoMax~(\emph{PIM}), including a curriculum negative sampling method and a path representation learning method. 
First, we propose a \emph{curriculum negative sampling strategy} to generate a small number of negative paths for an input path. 
Instead of randomly select other input paths as negative paths, the strategy follows the principles of curriculum learning~\cite{DBLP:conf/icml/BengioLCW09} to first generate paths that are largely different from the input path and thus are easy to be distinguished from the input path. Then, we gradually generate paths that are increasingly similar to the input path and thus are more difficult to be distinguished from the input path.  
The proposed curriculum negative sampling facilitates effective learning of distinguishable path representations. 

Next, we propose two different discriminators, a \emph{path-path discriminator} and a \emph{path-node discriminator}, to jointly learn path representations. 
The path-path discriminator captures the representation differences between an input path and its negative paths, which we refer to as a \emph{global} view. %
The path-node discriminator captures the representation difference between an input path and the representations of the nodes that only appear in its negative paths, which we refer to as a \emph{local} view. %
The two discriminators ensure the quality of the learned path representations, because they are distinguishable from not only the representations of negative paths from a global view but also the representations of the nodes in negative paths from a local view. %
To the best of our knowledge, \emph{PIM} is the first work that studies unsupervised path representation learning. First, we propose a curriculum negative sampling strategy for path representation learning. Second, we propose the path-path and path-node discriminators to learn jointly path representations from a global and a local view. Third, we conduct extensive experiments on two data sets with two downstream tasks to demonstrate the effectiveness of \emph{PIM}. We make the following contributions. 
\begin{enumerate}
  \item We propose a curriculum negative sampling strategy for path representation learning.
  \item We propose the path-path and path-node discriminators to learn jointly path representations from a global and a local view.
  \item We conduct extensive experiments on two data sets with two downstream tasks to demonstrate the effectiveness of \emph{PIM}.
\end{enumerate}

\section{Related Work}
\noindent
\subsection{Path Representation Learning.}
Existing proposals on path representation learning are all under the supervised learning setting. Such proposals often require large amount of labeled training data and the learned path representations cannot be easily reused for other tasks. 
For example, \emph{Deepcas}~\cite{DBLP:conf/www/LiMGM17}, \emph{ProxEmbed}~\cite{DBLP:conf/aaai/LiuZZZCWY17}, and \emph{PathRank}~\cite{yang2020context,DBLP:conf/icde/Yang020} employ different kinds of RNNs to combine node representations of the nodes in a path to obtain a path representation. Then, the training is performed in an end-to-end fashion by using the labeled training data. 
%
%
Instead, we propose an unsupervised path representation learning framework \emph{PIM} that does not require labeled training data and it generalizes nicely to multiple downstream tasks (cf. Table~\ref{tb1:ttpr} in Section~\ref{sssec:overallaccuracy}). In addition, \emph{PIM} can be used as a pre-training method to enhance existing supervised path representation learning (cf. Figure~\ref{fig:pre-tt-pr} in Section~\ref{sssec:pretrain}). 
An unsupervised trajectory representation learning method transforms trajectories into images and thus do not apply on paths in graphs~\cite{DBLP:conf/cikm/Kieu0GJ18}.

\noindent
\subsection{Mutual Information Maximization on Graphs.}
Motivated by Deep InfoMax~\cite{DBLP:conf/iclr/HjelmFLGBTB19}, mutual information maximization has been applied for unsupervised graph representation learning. 
Deep Graph Infomax~(\emph{DGI})~\cite{DBLP:conf/iclr/VelickovicFHLBH19} and Graph Mutual Information~(\emph{GMI})~\cite{DBLP:conf/www/PengHLZRXH20} learn node representations and \emph{InfoGraph}~\cite{DBLP:conf/iclr/SunHV020} learns whole graph representations. 
Here, negative samples are often randomly drawn from a different graph and the mutual information only considers a local view, e.g., a node representation vs. a graph representation. 
In \emph{PIM}, we propose a curriculum negative sample strategy to generate negative paths with different overlapping nodes with the input paths from the same graph, which facilitates training. 
Other advanced negative sampling approaches exist~\cite{DBLP:conf/aaai/WangLP18,DBLP:conf/nips/DingQY0J20}, but they are not proposed for graphs and do not follow curriculum learning. 
In addition, we compute mutual information on both a local view (i.e., the representations of input paths vs. the node representations of negative paths) and a global view (i.e., the representations of input paths vs. negative paths) and use them jointly to train the model, which improves accuracy.   
%

%
\begin{figure*}[t]
\centering
\includegraphics[width=0.9\textwidth]{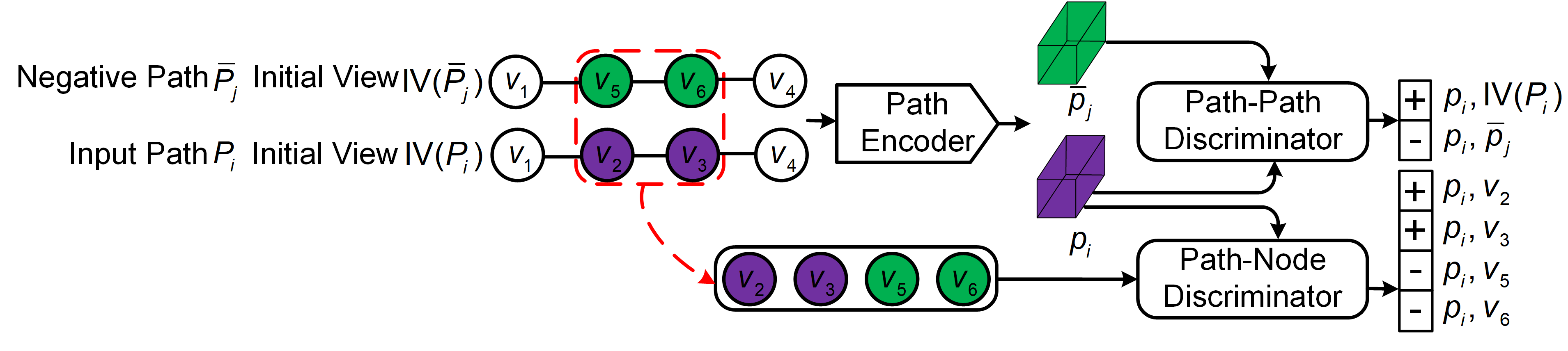}
  \caption{\emph{PIM} Overview. The Path Encoder takes as input the initial view $IV(P_i)$ of input path $P_i$ and the initial view $IV(\bar{P}_j)$ of negative path $\bar{P}_j$, and returns their representations $p_i$ and $\bar{p}_j$, respectively. 
  The Path-Path Discriminator takes as input a pair of path representations and decides whether they are from the same path. A positive pair, e.g., $(p_i, IV(P_i))$, refers to two different representation views of the same input path $P_i$. A negative pair, e.g., $(p_i, \bar{p}_j)$, refers to the path representations of an input path vs. its negative path.  
  The Path-Node Discriminator takes as input a (input path representation, node feature vector) pair and decides whether the node is from the input path. 
  A positive pair, e.g., $(p_i, v_2)$, represents the path representation of $P_i$ and a node feature vector of node $v_2$ that only appears in $P_i$. A negative pair, e.g., $(p_i, v_5)$, represents the path representation of the input path and a node feature vector of node $v_5$ that only appears in the negative path.
  }
  \label{fig:Framework}
\end{figure*}

section{Preliminaries}
\label{sec:prelim}

\noindent
\paragraph{Graph.} We consider a directed graph $G=(\mathbb{V}, \mathbb{E})$, where $\mathbb{V}$ is the node set and $\mathbb{E}$ is the edge set and we have $|\mathbb{V}|=N$ and $|\mathbb{E}|=M$. 
Each node $V_i\in \mathbb{V}$ is associated with a node feature vector $v_i\in \mathbb{R}^{D}$.  

\noindent
\paragraph{Path.} A path $P=\langle V_1, V_2, \ldots, V_Z \rangle$ is a sequence of nodes, where $Z$ is the path length and $P.s=V_1$ and $P.d=V_Z$ are the source and destination of path $P$, respectively. Each pair of adjacent nodes $(V_k$, $V_{k+1})$ is connected by an edge in $\mathbb{E}$, $1\leq k< Z$. 
We use $IV(P)\in \mathbb{R}^{Z\times D}$ to represent the concatenation of the node feature vectors of the nodes in path $P$. 
We call $IV(P_i)$ the initial view of path $P_i$. 

\noindent
\paragraph{Problem Definition.} 
Given a set of path $\mathbb{P}$ in graph $G$, \emph{Path Representation Learning (PRL)} aims at learning a path representation vector $p_i\in \mathbb{R}^{D^\prime}$ for each path $P_i\in \mathbb{P}$.  
Formally, \emph{PRL} learns a path encoder $PE_{\psi}$ that takes as input the initial view $IV(P_i)$ of path $P_i$, i.e., the node features of the nodes in path $P_i$, and outputs its path representation vector $p_i$. 
\begin{equation}
    PE_{\psi}: {\mathbb{R}}^{Z\times D} \rightarrow {\mathbb{R}}^{D'}, 
\end{equation}
\noindent
where $\psi$ indicates the learnable parameters for the path encoder, e.g., weights in a neural network, $Z$ is the length of path $P_i$, and $D'\ll Z\times D$ is an integer indicating the dimension of the learned path representation vector $p_i$.

The learned path representation vectors are supposed to support a variety of downstream tasks, e.g., path ranking and path travel time estimation. 
\section{Path InfoMax} 
Figure~\ref{fig:Framework} offers an overview of the proposed framework Path InfoMax (\emph{PIM}). %
\emph{PIM} employs contrastive learning, specifically mutual information maximization, to train the path encoder to produce path representations without requiring task-specific labels. %

The path encoder takes as input the initial view of an input path and outputs its path representation (cf. Sec.~\ref{ssec:PE}).
Training the path encoder is supported by a path-path discriminator and a path-node discriminator using negative samples. %
To this end, we first introduce the curriculum negative sampling strategy to generate negative paths (cf. Sec.~\ref{ssec:negsampling}). %
Then, the path-path discriminator guides the path encoder to produce path representations such that the path representations of input paths can be distinguished from the path representations of negative paths %
(cf. Sec.~\ref{ssec:GMIM}). 
In addition, the path-node discriminator guides the path encoder to produce path representations such that the path representations of input paths can be distinguished from the node features of the nodes that only appear in the negative paths 
(cf. Sec.~\ref{ssec:LMIM}). %
Finally, we discuss the final training objectives of~\emph{PIM}.

\subsection{Path Encoder} 
\label{ssec:PE}
Since a path consists of a sequence of nodes, we use a model that is able to encode sequential data, e.g., a recurrent neural network~\cite{DBLP:journals/neco/HochreiterS97,DBLP:conf/ssst/ChoMBB14} or a Transformer~\cite{DBLP:conf/nips/VaswaniSPUJGKP17} as the path encoder $PE_{\psi}$, where $\psi$ represents the parameters to be learned for the path encoder.

Given a path $P_i=\langle V_1, V_2, \ldots, V_Z \rangle$, we use its initial view $IV(P_i)\in \mathbb{R}^{Z\times D}$ as the input to the path encoder, which returns its path representation vector $p_i\in \mathbb{R}^{D'}$. 
\subsection{Curriculum Negative Sampling}
\label{ssec:negsampling}
\begin{figure}[t]
\centering
\includegraphics[width=0.7\columnwidth]{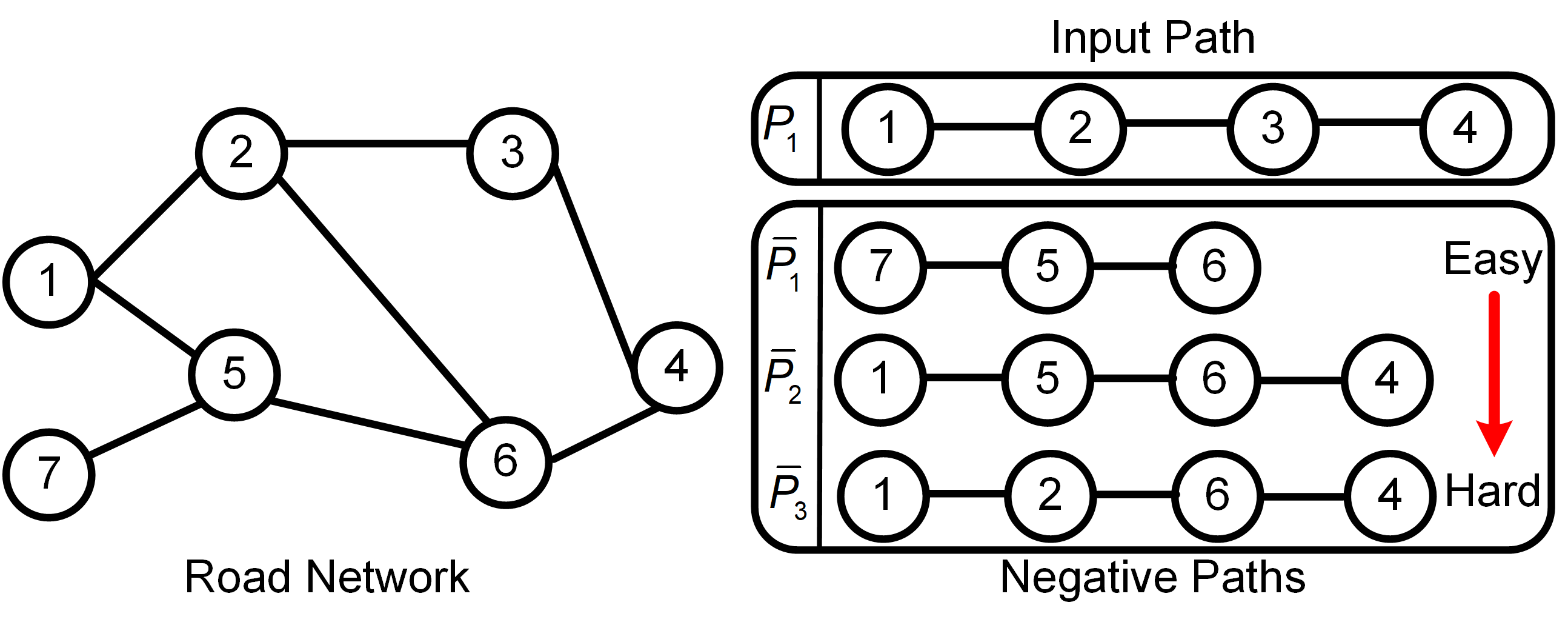}
  \caption{Curriculum Negative Sampling.}
  \label{fig:ns}
\end{figure}

Motivated by curriculum learning~\cite{DBLP:conf/icml/BengioLCW09}, we propose a curriculum negative sampling method to generate negative samples. 
The idea behind curriculum learning is that we start to train a model with easier samples first, and then gradually increase the difficulty levels. 
In our setting, we first generate negative paths that are different from the input path, e.g., paths without any overlapping nodes with the input path. In this case, it can be easy to train a path encoder that returns distinguishable representations of the input path and the negative paths. 
Then, we gradually generate negative paths that are increasingly similar to the input path, e.g., sharing the same source and destination with the input path and with increasingly overlapping nodes. 
This makes more difficult for the path encoder to generate distinguishable path representations. 
Figure~\ref{fig:ns} shows three negative paths $\bar{P_1}$, $\bar{P_2}$, and $\bar{P_3}$ with increasingly difficulties for input path $P_1$, along with the underlying road network graph. 

Specifically, for each input path $P_1$, we first randomly select a path from the path set $\mathbb{P}$ as the first negative path. Next, we use the source and the destination of $P_1$ as the input to call the top-k diversified shortest path algorithm~\cite{DBLP:journals/tkde/LiuJYZ18} to generate paths that share the same source and destination of $P_1$. This algorithm allows us to set different diversity thresholds, enabling us to generate negative paths with different overlapping nodes with the input path. 

\subsection{Global Mutual Information Maximization}
\label{ssec:GMIM}
We proceed to the learning of the path encoder using the negative paths. We first consider a global view of the path representations. We expect that the learned path representations are distinguishable from the path representations of the negative paths. 

To this end, we first construct negative and positive pairs for training a \emph{path-path discriminator} $D^{PP}_{\omega_1}$. 
In a negative pair $\langle(p_i, \bar{p}_j), -\rangle$, $p_i$ and $\bar{p}_j$ represent the path representations of input path $P_i$ and a negative path $\bar{P}_j$, respectively, which are both returned by the path encoder $PE_{\psi}$. 
In a positive pair $\langle(p_i, IV(P_i), +\rangle$, $p_i$ is still the path representations of input path $P_i$ returned by the path encoder and $IV(P_i)$ is the initial view of path $P_i$ (cf. Section~\ref{ssec:PE}). Here, $p_i$ and $IV(P_i)$ represent two different views, i.e., a view from the path encoder vs. a view from the node features, of the same input path $P_i$. Figure~\ref{fig:Framework} shows examples of a negative and a positive pair.  

Next, we use mutual information maximization to train the path-path discriminator $D^{PP}_{\omega_1}$ such that it is able to make a binary classification on the negative vs. positive pairs. 
Specifically, we aim at maximizing the estimated mutual information (MI) over the positive and negative pairs. 
$$
    \argmax_{\psi, \omega_1} \sum_{P_i\in \mathbb{P}} I_{\psi, \omega_1} (p_i, \mathbb{NP}_i),
$$
where $I_{\psi, \omega_1}(\cdot, \cdot)$ is the MI estimator modeled by the path-path discriminator $D^{PP}_{\omega_1}$ that is parameterized by parameters $\omega_1$ and the path encoder $PE_\psi$ that is parameterized by parameters $\psi$. Path $P_i$ is an input path from $\mathbb{P}$ and $p_i$ is its path representation returned by the path encoder. %
$\mathbb{NP}_i$ includes the negative paths of $P_i$. 
Following~\cite{DBLP:conf/iclr/VelickovicFHLBH19,DBLP:conf/iclr/HjelmFLGBTB19}, we use a noise-contrastive type objective with a standard binary cross-entropy loss on the positive pairs and the negative pairs, as shown in Equation~\ref{eq:bcemi}. 
\begin{align}
\small
\begin{split}
\label{eq:bcemi}
\mathcal{I}_{\psi, \omega_1}\left(p_i, \mathbb{NP}_{i}\right):=& \frac{1}{1+|\mathbb{NP}_{i}|}( \mathbb{E}_{\mathbb{P}}\left[\log {D}^{PP}_{\omega_1}\left(p_{i},IV(P_i)\right)\right]+ \\
    & \sum_{\bar{P}_j\in \mathbb{NP}_{i}}
    \mathbb{E}_{\mathbb{NP}_i}\left[\log \left(1-D^{PP}_{\omega_1}\left(p_{i}, \bar{p}_{j}\right)\right)\right])
\end{split}
\end{align}
\noindent
Here, we use $\mathbb{E}_\mathbb{P}$ and $\mathbb{E}_{\mathbb{NP}_i}$ to denote expectations w.r.t. the empirical probability distribution of the input paths and the negative paths. 
Note that $p_i$ and $\bar{p}_j$ are the path representations returned by the path encoder $PE_{\psi}$. Thus, maximizing the MI estimator enables the training of both the path encoder (i.e., parameters $\psi$) and the path-path discriminator (i.e., parameters $\omega_1$). 
\subsection{Local Mutual Information Maximization}
\label{ssec:LMIM}

We proceed to consider a local view of the path representations. We expect that the learned path representations are distinguishable from the node feature vectors of the nodes from input vs. negative paths. This is particularly important when distinguishing two paths with significant overlapping nodes. 
We introduce a positive node set $\mathbb{X}_i$ that includes nodes appearing only in the input path $P_i$ but not the negative paths and a negative node set $\mathbb{Y}_i$ that includes nodes appearing only in the negative paths but not the input path $P_i$. 
We then construct negative and positive pairs for training a \emph{path-node discriminator} $D^{PN}_{\omega_2}$. 
In a negative pair $\langle(p_i,{v}_j), -\rangle$, $p_i$ represents the path representations of input path $P_i$, returned by the path encoder $PE_{\psi}$; ${v}_j$ represents the node feature vector of a negative node $V_j\in \mathbb{Y}_i$.  
Similarly, in a positive pair $\langle(p_i, v_k), +\rangle$, ${v}_k$ represents the node feature vector of a positive node $V_k\in \mathbb{X}_i$.  Figure~\ref{fig:Framework} shows examples of two negative and two positive such pairs for the path-node discriminator. 

Similar to the path-path discriminator training, we also employ mutual information maximization to train the path-node discriminator $D^{PN}_{\omega_2}$. 
In particular, we have 
$$\argmax_{\psi, \omega_2} \sum_{P_i\in \mathbb{P}} 
I_{\psi, \omega_2} (p_i, \mathbb{X}_i\cup\mathbb{Y}_i),$$
where $I_{\psi, \omega_2}$ is the MI estimator modeled by the path-node discriminator $D^{PN}_{\omega_2}$ that is parameterized by parameters $\omega_2$ and the path encoder $PE_\psi$ that is parameterized by parameters $\psi$. We use a noise-contrastive with a BCE loss, similar to Equation~\ref{eq:bcemi}, to compute $I_{\psi, \omega_2}(p_i, \mathbb{X}\cup\mathbb{Y})$ as follows. 

\begin{align}
\small
\begin{split}
\label{eq:bcemiPN}
I_{\psi, \omega_2} (p_i, \mathbb{X}_i\cup\mathbb{Y}_i):=& \frac{1}{|\mathbb{X}_i\cup\mathbb{Y}_i|}
(
\sum_{v_k\in \mathbb{X}_{i}}\mathbb{E}_{\mathbb{X}_i}\left[\log \mathcal{D}^{PN}_{\omega_2}\left(p_{i},v_k\right)\right]+ \\
    & \sum_{{v}_j\in \mathbb{Y}_{i}}
    \mathbb{E}_{\mathbb{Y}_i}\left[\log \left(1-D^{PN}_{\omega_2}\left(p_{i}, {v}_{j}\right)\right)\right] )
\end{split}
\end{align}
\subsection{Maximization of \emph{PIM}}

We combine both the global and local mutual information maximization when training the final \emph{PIM} model, see below. 
$$\argmax_{\psi, \omega_1,  \omega_2} \sum_{P_i\in \mathbb{P}} \left( I_{\psi, \omega_1} (p_i, \mathbb{NP}_i) +  I_{\psi, \omega_2} (p_i, \mathbb{X}_i\cup\mathbb{Y}_i) \right).$$
\section{Experiments}
We conduct experiments to investigate the effectiveness of the proposed unsupervised path representation learning framework \emph{PIM} on two downstream tasks using two data sets. In addition, we also demonstrate that \emph{PIM} is able to use as a pre-training method to enhance supervised path representation learning.  

\subsection{Experimental Setup}

\subsubsection{Road Network and Paths} 
We obtain two road network graphs from OpenStreetMap. The first is from Aalborg, Denmark, consisting of 8,893 nodes and 10,045 edges. The second is from Harbin, China, consisting of 5,796 nodes and 8,498 edges. We also obtain two substantial GPS trajectory data sets on the two road networks.  %
We consider 52,494 paths in the Aalborg network and 37,079 paths in the Harbin network. 
\subsubsection{Downstream Tasks} 
%
    \paragraph{Path Travel Time Estimation.}Each path is associated with its travel time (seconds) obtained from trajectories. We aim at building a regression model to estimate the travel time of paths. We evaluate the accuracy of the estimations by Mean Absolute Error (\textbf{MAE}), Mean Absolute Relative Error (\textbf{MARE}) and Mean Absolute Percentage Error (\textbf{MAPE}). 
    
    \paragraph{Path Ranking.}Given a set of paths, which often share the same source and destination, each path is associated with a ranking score in range $[0, 1]$. The ranking scores are obtained with the help of trajectories by following an existing study~\cite{yang2020context}. In path ranking, we aim at building a regression model to estimate the ranking scores of the paths. To evaluate the accuracy of the estimated ranking scores, we not only report the \textbf{MAE} of the estimated ranking scores but also use Kendall rank correlation coefficient (denoted by $\tau$) and Spearman's rank correlation coefficient (denoted by $\rho$) to measure the consistency between the ranking derived by the estimated ranking scores vs. the ranking derived by the ground truth ranking scores. Smaller \textbf{MAE} and higher $\tau$ and $\rho$ values indicate higher accuracy. 


%

\subsubsection{Baselines}
We compare \emph{PIM} with seven baseline methods. 
\begin{itemize}
    \item \textbf{Node2vec}~\cite{DBLP:conf/kdd/GroverL16}, Deep Graph InfoMax (\textbf{DGI})~\cite{DBLP:conf/iclr/VelickovicFHLBH19}, Graphical Mutual Information Maximization (\textbf{GMI})~\cite{DBLP:conf/www/PengHLZRXH20} are three unsupervised node representation learning models, which output the node representation for each node in a graph. We use the average of the node representations of the nodes in a path to get the path representation of the path. We also consider using concatenation instead of average, but resulting worse accuracy.
    \item \textbf{Memory Bank (\textbf{MB})}~\cite{DBLP:conf/cvpr/WuXYL18} is an unsupervised learning approach to obtain representations from high-dimensional data. It uses a memory bank to achieve the negative samples from current batch to train an encoder, then gets the representation based on contrastive loss.
    We re-implement \textbf{MB} with an LSTM encoder to better capture the sequential information to get the path representations.   
    \item \textbf{InfoGraph}~\cite{DBLP:conf/iclr/SunHV020} is an unsupervised whole graph representation learning model. Here, we treat a path as a graph to learn the path representation. 
    \item \textbf{BERT}~\cite{DBLP:conf/naacl/DevlinCLT19} is an unsupervised language representation learning model. To enable training, we (1) treat a path as a sentence and mask some nodes in the path; and (2) split a path $P$ into two sub-paths $P_1$ and $P_2$, and consider $(P_1, P_2)$ as a valid Q\&A pair and $(P_2, P_1)$ as an invalid Q\&A pair because the former keeps a meaningful order while the latter does not. 
    \item \textbf{PathRank}~\cite{yang2020context} is a \emph{supervised} path representation learning model based on GRU. \textbf{PathRank} takes into account the labels from a specific downstream task to obtain path representations.  
\end{itemize}

\noindent
Among these baselines, \emph{Node2vec}, \emph{DGI}, \emph{GMI}, \emph{MB}, \emph{InfoGraph}, and \emph{BERT} are \emph{unsupervised learning} approaches, which do not employ labels from specific downstream tasks to produce path representations. 
In contrast, \emph{PathRank} is a \emph{supervised learning} approach, where it employ labels from specific downstream tasks to produce path representations, meaning that the obtained path representations are different when using labels from different downstream tasks. 
\subsubsection{Regression Model}
For all unsupervised learning approaches, we first obtain a task-independent path representation and then apply a regression model to solve different downstream tasks using task-specific labels. %
In the experiments, we choose \textit{Gaussian Process Regressor} (\emph{GPR}) 
to make travel time and ranking score estimation. 
We randomly choose 85\%, 10\%, and 5\% of the paths as the training, validation, and test sets.  
\subsubsection{Implementation Details}
We use an LSTM as the path encoder. 
We use \emph{node2vec}~\cite{DBLP:conf/kdd/GroverL16}, an unsupervised node representation learning method, to obtain a 128 dimensional node feature vector for each node, i.e., $D=128$. 
We set the path representation size $D^\prime=128$. 
In the curriculum negative sampling, for each input path, we generate four negative paths---the first two paths are randomly selected from $\mathbb{P}$ and the third and the fourth paths are two paths returned by the top-k diversified shortest paths with different overlapping nodes with the input path. 
We use Adam~\cite{DBLP:journals/corr/KingmaB14} for optimization with learning rate of 0.001. All algorithms are implemented in Pytorch 1.7.1. We conduct experiments on Ubuntu 18.04.5 LTS, with 40 Intel(R) Xeon(R) Gold 5215 CPUs @ 2.50GHz and four Quadro RTX 8000 GPU cards. Code is available at  \url{https://github.com/Sean-Bin-Yang/Path-InfoMax.git}.
\subsection{Experimental Results}
\label{sec:ER}

\begin{table*}[t]
\small
\centering

\begin{tabular}{l|llllll|llllll}
\toprule[2pt]
\multirow{3}{*}{\textbf{Method}} & \multicolumn{6}{l|}{\textbf{Aalborg}}                                                     & \multicolumn{6}{l}{\textbf{Harbin}}                                                     \\ \cline{2-13} 
                        & \multicolumn{3}{l|}{\textbf{Travel Time Estimation}}  & \multicolumn{3}{l|}{\textbf{Path Ranking}} & \multicolumn{3}{l|}{\textbf{Travel Time Estimation}} & \multicolumn{3}{l}{\textbf{Path Ranking}} \\ \cline{2-13} 
                        &\textbf{MAE} &\textbf{MARE} &\textbf{MAPE} & \textbf{MAE}    &\textbf{$\tau$} &\textbf{$\rho$}    &\textbf{MAE}     & \textbf{MARE}    & \textbf{MAPE}   & \textbf{MAE} &\textbf{$\tau$} &\textbf{$\rho$} \\ \toprule[1pt]
\emph{Node2vec}   & 121.43    & 0.27   & 31.04      &0.18        & 0.66           & 0.70               & 258.91    & 0.22  & 23.17       &0.15      &0.70                 &0.72                 \\ \hline
\emph{DGI}        & 192.63    & 0.42   & 82.44      &0.54        & 0.49           & 0.52               & 528.71    & 0.39  & 86.53       &0.21       &0.59                 &0.60                  \\ \hline
\emph{GMI}        & 136.58    & 0.30   & 50.81      &0.23        & 0.58           & 0.61               & 979.68    & 0.73  & 192.45      &0.24       &0.55                 &0.56                  \\ \hline
\emph{MB}         & 243.97    & 0.53   & 84.17      &0.35        & 0.34           & 0.38               & 533.41    & 0.40  & 86.01       &0.27       &0.31                 &0.34                \\ \hline
\emph{BERT}       &254.17     &0.54    &61.61       &0.36        &0.38            &0.39                &514.95     &0.57   & 49.80       &0.28       &0.45                 &0.46                \\ \hline
\emph{InfoGraph}  &132.28     &0.29    &39.47       &0.17        &0.69           &0.73              &391.45      &0.44   &44.60          &0.29       &0.68                 & 0.72                \\ \hline
\emph{PIM}   &\textbf{76.10}     &\textbf{0.16}      &\textbf{17.28}      &\textbf{0.12}  &\textbf{0.72}     &\textbf{0.76}                & \textbf{125.76}    & \textbf{0.14}  & \textbf{13.73}         &\textbf{0.11}      &\textbf{0.75}                 &\textbf{0.79}                 \\ \toprule[2pt]
\end{tabular}
\caption{{Overall Accuracy on Travel Time Estimation and Ranking Score Estimation.}}
\label{tb1:ttpr}
\end{table*}
\subsubsection{Overall accuracy on both downstream tasks}
\label{sssec:overallaccuracy}

Table~\ref{tb1:ttpr} shows the results on travel time and ranking score estimation. 
\emph{PIM} consistently outperforms all baselines on both tasks and on both data sets. 
\emph{Node2vec}, \emph{DGI}, and \emph{GMI} fail to capture the dependencies among node feature vectors in paths. In contrast, \emph{PIM} considers such dependencies by using the LSTM based path encoder. In addition, the two discriminators further improve the accuracy. 

\emph{InfoGraph} implicitly considers node feature vector sequences. However, the discriminator in \emph{InfoGraph} only considers the local view. In addition, \emph{InfoGraph} considers other paths in the same batch as negative samples, whereas \emph{PIM} employs curriculum negative sampling to generate negative samples. \emph{PIM} outperforms \emph{InfoGraph} suggests that the proposed curriculum negative sampling and jointly consider both local and global views are effective.  

Although \emph{MB} and \emph{BERT} also capture dependencies among the node feature vectors in paths, such methods only achieve %
relatively poor accuracy. This is because \emph{MB} often requires large amount of negative samples (e.g., more than 256), which is not feasible in our setting. Although the unsupervised training strategy in \emph{BERT} works well for NLP, it does not fit our problem setting on learning path representations. 


\subsubsection{Using \emph{PIM} as a Pre-training Method}
\label{sssec:pretrain}
In this experiment, we consider \emph{PIM} as a pre-training method for the supervised method \emph{PathRank}. 
\emph{PathRank} employs an GRU that takes as input node feature vectors in a path and predicts travel time or ranking scores. 
To use \emph{PIM} as a pre-training method for \emph{PathRank}, we use a GRU based path encoder. Then, we first train \emph{PIM} in an unsupervsied manner, and then use the learned parameters in the GRU path encoder to initialize the GRU in \emph{PathRank}. Finally, we use the labelled training paths to fine tune \emph{PathRank}. 

\begin{figure}
\centering    
\subfigure[Travel Time Estimation] {
 \label{fig:a}     
\includegraphics[ width=0.35\columnwidth]{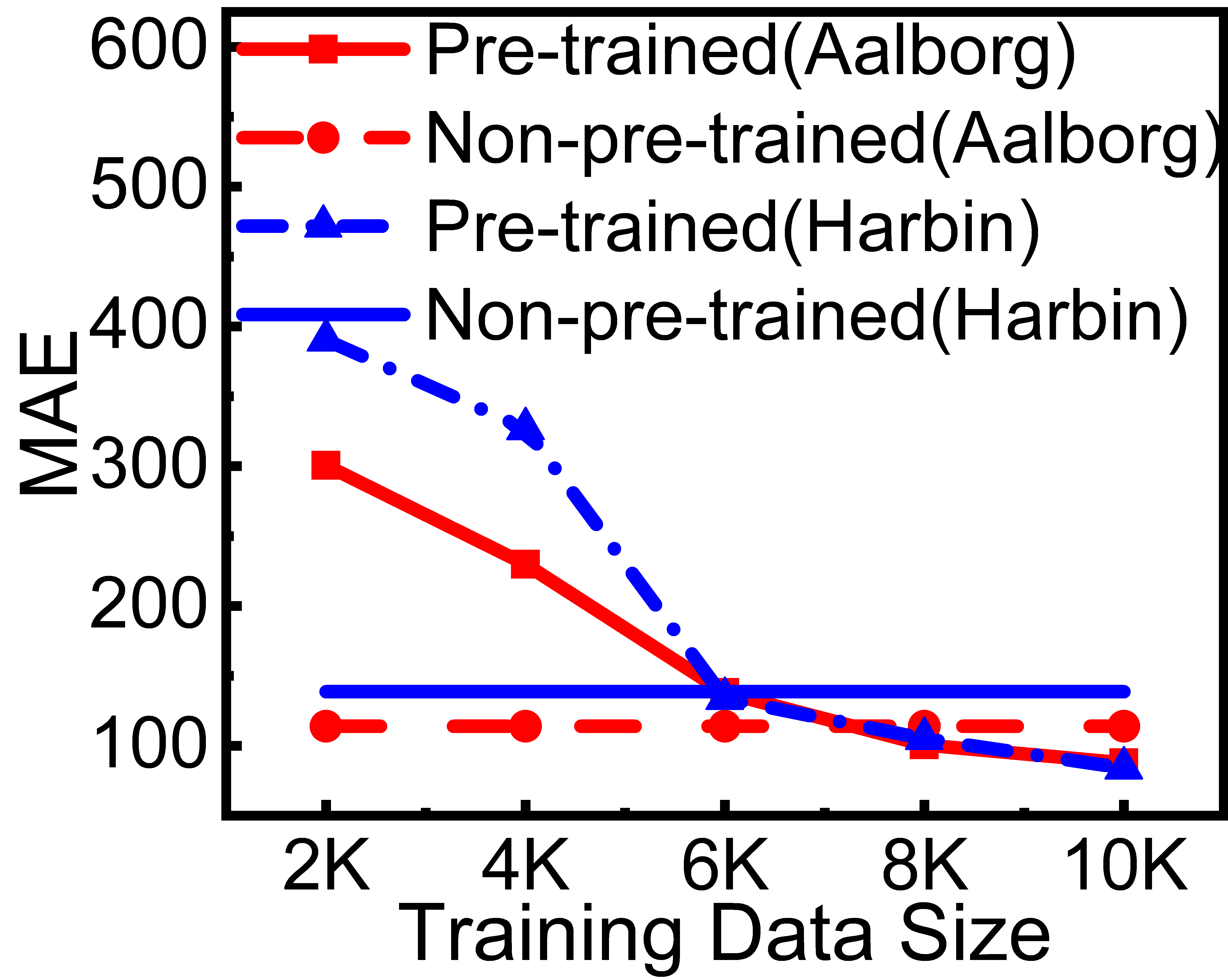}  
}     
\subfigure[Path Ranking] { 
\label{fig:b}     
\includegraphics[width=0.35\columnwidth]{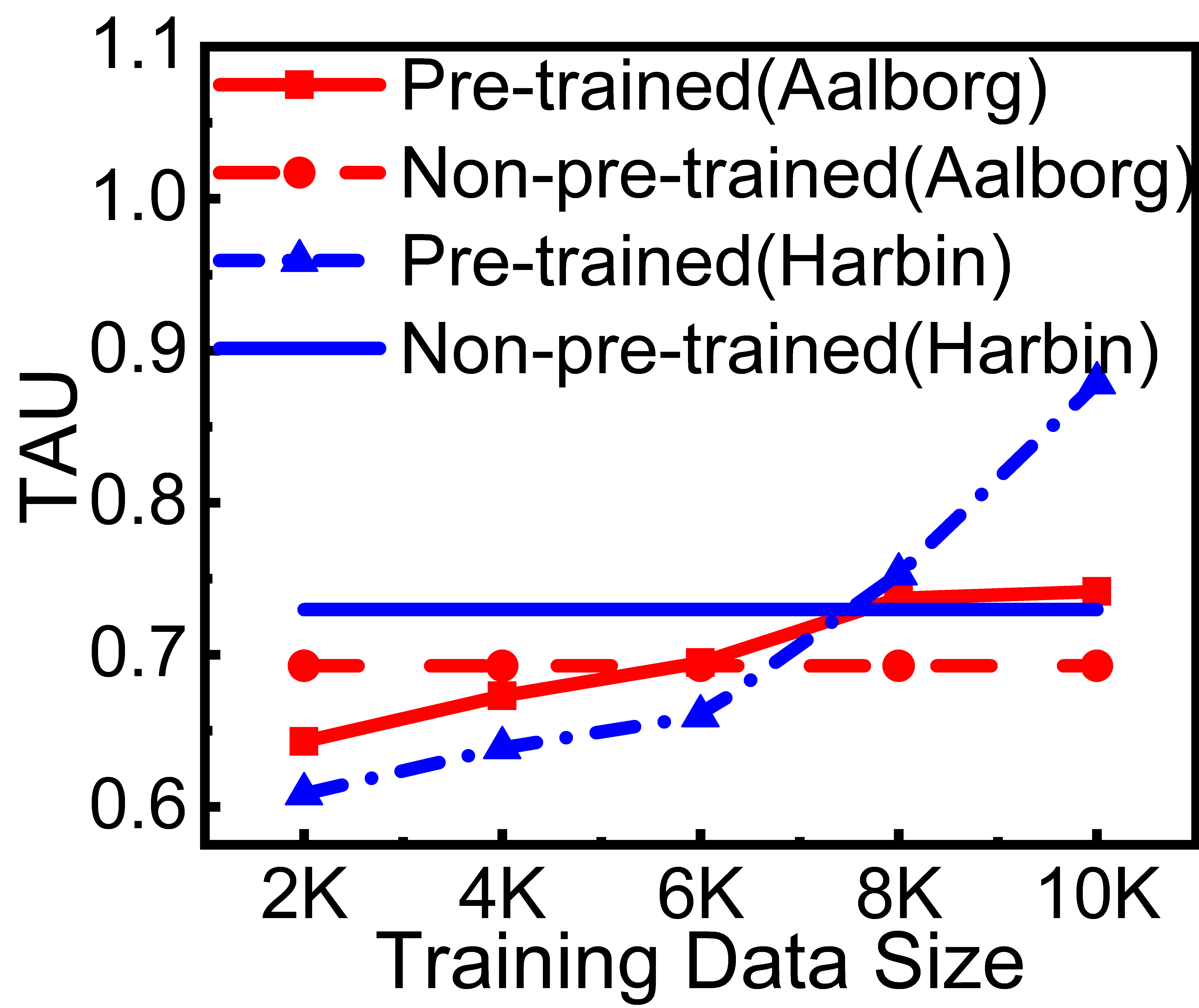}     
}    
\caption{Effects of Pre-training.}     
\label{fig:pre-tt-pr}     
\end{figure}

Figure~\ref{fig:pre-tt-pr} shows the travel time estimation performance of \emph{PathRank} with vs. without pre-training on both data sets. When not using pre-training, we train \emph{PathRank} using 10K labelled training paths.  
We observe that: (1) when using pre-training, we are able to achieve the same accuracy of the non-pre-training \emph{PathRank} using less labelled training paths, e.g., ca. 7K for Aalborg and 6K for Harbin. 
(2) when using 10K labelled training paths, the pre-training \emph{PathRank} achieves higher accuracy than the non-pre-training \emph{PathRank}. 
We observe similar results on the other task of path ranking, suggesting that \emph{PIM} can be used as a pre-training method to enhance supervised methods.

\subsection{Ablation Studies}
\noindent

\noindent
\subsubsection{Impact of Local and Global MI Maximization} 
%
We investigate the impact of jointly using both path-path and path-node discriminators to consider both the local and global MI maximization. We consider two variants of \emph{PIM} where (1) we only use the path-path discriminator to maximize the global MI and (2) we only use the path-node discriminator to maximize the local MI. 
Table~\ref{tb:np} shows that jointly maximizing both the local and global MI achieves the best accuracy, which justifies our design choices of using both the path-path and path-node discriminators. 

\begin{table}[]
\small
\centering


\begin{tabular}{l|llllll}
\toprule[2pt]
\multirow{2}{*}{} & \multicolumn{3}{l}{\textbf{Travel Time Estimation}} & \multicolumn{3}{l}{\textbf{Path Ranking}} \\ \cline{2-7}
                  &\textbf{MAE} &\textbf{MARE} &\textbf{MAPE} & \textbf{MAE}    &\textbf{$\tau$} &\textbf{$\rho$} \\ \toprule[1pt]
\textbf{Global} &237.92     &0.51      &85.88     &0.34  &0.22     &0.25     \\ 
\textbf{Local}  &{118.03}     &{0.25}      &{26.20}     &0.14  &{0.70}     &{0.74} \\ 
\textbf{Joint}  &\textbf{76.10}     &\textbf{0.16}      &\textbf{17.28}      &\textbf{0.12}  &\textbf{0.72}     &\textbf{0.76}    \\ \toprule[2pt] 
\end{tabular}
\caption{Effects of Local and Global MI Maximization, Aalborg.}
\label{tb:np}
\end{table}

\noindent
\subsubsection{Impact of Curriculum Negative Sample Strategy}
To investigate the effectiveness of the proposed curriculum negative sample strategy, we compare it with the following two strategies. 

 \begin{enumerate}
\item Random only: it randomly selects paths from $\mathbb{P}$. 
\item Top-k only: it employs the top-k diversified shortest path algorithms to generate  negative paths sharing the same origin and destination with the input path with different overlapping nodes. 

\end{enumerate}
To make a fair comparison, we use each strategy to generate the same number of negative paths, i.e., 4. 
Table~\ref{tb1:cn} shows that the top-k only strategy is better than random only, suggesting that it is important to distinguish the representations of input paths vs. paths sharing the same origin and destination. 
The proposed curriculum negative sampling strategy achieves the best accuracy, suggesting that training \emph{PIM} from easy to hard negative paths help further improves accuracy. 

\begin{table}[t]
\small
\centering

\begin{tabular}{l|lll|lll}
\toprule[2pt]
\multirow{2}{*}{\begin{tabular}[c]{@{}l@{}}\textbf{} \\ \textbf{}\end{tabular}} & \multicolumn{3}{l|}{\textbf{Travel Time Estimation}} & \multicolumn{3}{l}{\textbf{Path Ranking}} \\ \cline{2-7}
                                                                             & \textbf{MAE} & \textbf{MARE} & \textbf{MAPE}     &\textbf{MAE}     &\textbf{$\tau$} &\textbf{$\rho$} \\ \toprule[1pt]
\textbf{Rand.}                                                                            &101.16      &0.22       &23.51   &0.14    &0.65     &0.69          \\
\textbf{Top-k}                                                                            &100.87      &0.22      &22.31    &0.13    &{0.72}     &{0.75}                \\ 
\textbf{Curr.}       &\textbf{76.10}     &\textbf{0.16}      &\textbf{17.28}      &\textbf{0.12}  &\textbf{0.72}     &\textbf{0.76}\\  \toprule[2pt]
\end{tabular}
\caption{Effects of Curriculum Negative Sample Strategy, Aalborg.}
\label{tb1:cn}

\end{table}

\begin{table}[t]
\centering
\vspace{-5pt}
\label{tb:np}
\begin{tabular}{l|lll|lll}
\toprule[2pt]
\multirow{2}{*}{} & \multicolumn{3}{l|}{\textbf{Travel Time Estimation}} & \multicolumn{3}{l}{\textbf{Path Ranking}} \\ \cline{2-7}
                  &\textbf{MAE} &\textbf{MARE} &\textbf{MAPE} & \textbf{MAE}    &\textbf{$\tau$} &\textbf{$\rho$} \\ \toprule[1pt]
\textbf{K=1} &119.77     &0.29     &32.91     &0.19  &0.58     &0.63     \\ 
\textbf{K=2} &107.46      &0.26       &29.22   &0.18    &0.59     &0.63    \\ 
\textbf{K=3} &87.58      &0.19      &20.00     &0.12  &0.71     &0.74     \\ 
\textbf{K=4} &\textbf{76.10}     &\textbf{0.16}      &\textbf{17.28}      &\textbf{0.12}  &\textbf{0.72}     &\textbf{0.76}\\ \toprule[2pt]       
\end{tabular}
\caption{Effects of Negative Path Numbers, Aalborg.}
\label{tb:np}
\end{table}

\begin{table}[!h]
\centering
\begin{tabular}{l|lll|lll}
\toprule[2pt]
\multirow{2}{*}{\textbf{\begin{tabular}[c]{@{}l@{}}Posi. \\ Nods.\end{tabular}}} & \multicolumn{3}{l|}{\textbf{Travel Time Estimation}} & \multicolumn{3}{l}{\textbf{Path Ranking}} \\ \cline{2-7} 
                                      &\textbf{MAE} &\textbf{MARE} &\textbf{MAPE} & \textbf{MAE}    &\textbf{$\tau$} &\textbf{$\rho$} \\ \toprule[1pt] 
20\%                 & 114.31       & 0.25          & 24.92                &0.20              &0.65              &0.70       \\ 
40\%                 & 111.33       & 0.24          & 24.08                &0.16              &0.66              &0.70       \\ 
60\%                 & 104.57       & 0.23          & 22.94                  &0.14              &0.68              &0.71       \\ 
80\%                 & 101.31       & 0.23          & 22.56                 &0.13              &0.68              &0.72      \\ 
100\%                &\textbf{76.10}     &\textbf{0.16}      &\textbf{17.28}      &\textbf{0.12}  &\textbf{0.72}     &\textbf{0.76}      \\ \toprule[2pt]
\multirow{2}{*}{\textbf{\begin{tabular}[c]{@{}l@{}}Neg. \\ Nods.\end{tabular}}}  & \multicolumn{3}{l|}{\textbf{Travel Time Estimation}} & \multicolumn{3}{l}{\textbf{Path Ranking}} \\ \cline{2-7} 
                                      &\textbf{MAE} &\textbf{MARE} &\textbf{MAPE} & \textbf{MAE}    &\textbf{$\tau$} &\textbf{$\rho$} \\ \toprule[1pt]
20\%                  & 130.90       & 0.29          & 28.21                &0.19              &0.60              &0.65    \\ 
40\%                  & 110.86       & 0.24          & 25.30                 &0.15              &0.67              &0.70 \\
60\%                  & 105.70       & 0.23          & 24.01               &0.13              &0.67              &0.71              \\
80\%                  & 102.80       & 0.22          & 23.35                  &0.13              &0.68              &0.72              \\
100\%                  &\textbf{76.10}     &\textbf{0.16}      &\textbf{17.28}      &\textbf{0.12}  &\textbf{0.72}     &\textbf{0.76}              \\ \toprule[2pt]
\end{tabular}
\caption{Effects of Positive / Negative Nodes, Aalborg.}
\label{tb1:pn}
\end{table}

\noindent
\subsubsection{Impact of Negative Path Numbers}
%
We investigate the impact of using different numbers of negative paths. 
We vary the number of negative paths $K$ from 1, 2, 3, to 4. 
Recall that when using curriculum negative sampling, the first two paths are from $\mathbb{P}$ and the last two paths are from the top-k diversified shortest path finding algorithm. %
Table~\ref{tb:np} suggests that when using more negative paths, the accuracy improves. The accuracy improvements from 2 to 3 is the largest, suggesting that the top-k algorithm is very effective on generating high quality negative paths.

\noindent
\subsubsection{Impact of Positive/Negative Nodes in local MI} 
To study the impact of positive and negative nodes, we consider cases where we only use 20\%, 40\%, 60\%, 80\% of positive or negative nodes. 
Table~\ref{tb1:pn} shows that the accuracy increases when using more less positive and negative nodes. 

\subsubsection{Impact of Travel Distances}

We now study the effect of travel distance on the performance of different models for Aalborg dataset. To this end, we group the paths into subgroups by their distances (km) [0,5),[5,10),[10,15),($\geq$15), and investigate the performance of different models on each  subgroup. 

\begin{figure}
\centering    

\subfigure[Travel Time Estimation] {
 \label{fig:tda}     
\includegraphics[ width=0.35\columnwidth]{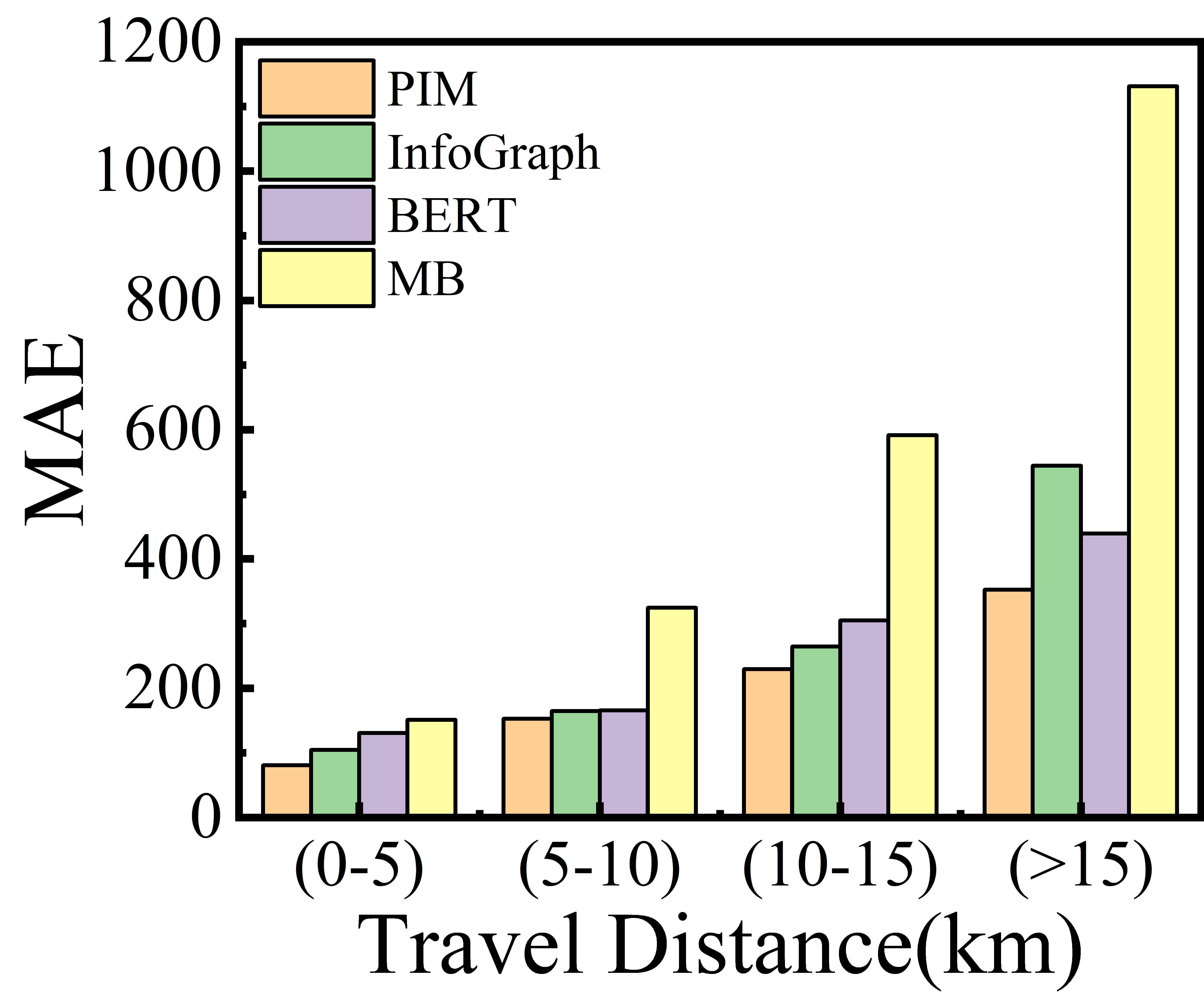}  
}     
\subfigure[Path Ranking] { 
\label{fig:tdb}     
\includegraphics[width=0.35\columnwidth]{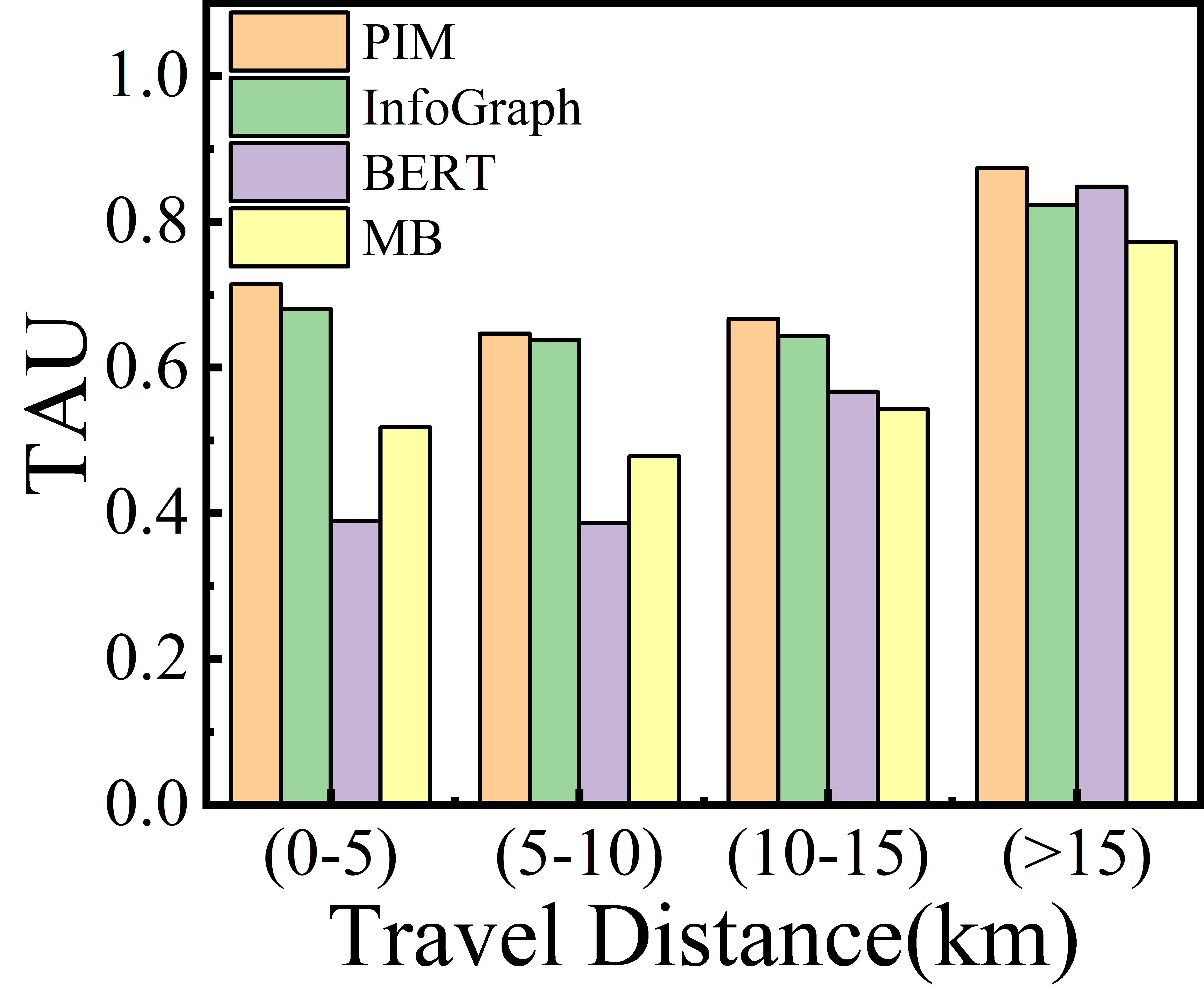}     
}    
\caption{Impact on Travel Distance.}     
\label{fig:td}     
\end{figure}

Figure~\ref{fig:td} plots MAEs and  Kendall rank correlation coefficient ($\tau$)  of \emph{InfoGraph}, \emph{BERT}, \emph{MB} and \emph{PIM} w.r.t. travel distance. we omit other baselines due to they can not directly achieve the path representation, which get the path representation through average all node's feature. 
Not surprisingly, MAEs, as shown in Fig~\ref{fig:tda}, increase with the distance since longer trips typically involve more road segments and have larger uncertainty and limited training data set. It is worth noting that the performance difference between \emph{PIM} and other baseline methods grows with the travel distance. This result suggests that the \emph{PIM} is more robust for estimating long paths' travel time. 
In addition, we observe that \emph{InfoGraph} and \emph{MB}, which both use contrastive learning, deteriorate as distance increases, suggesting that they are not capable of estimating travel time for long paths.
This is because that we have have less training data for long paths. This influences 
\emph{InfoGraph} and \emph{MB} efficiently since these two methods are sensitive to number of negative samples.
While for Path Ranking, as shown in Fig~\ref{fig:tdb}, it does not follow the tendency of MAEs, but our \emph{PIM} still achieves the best results on each distance subgroups.

\subsubsection{Computational performance of \emph{PIM}}

The learned \emph{PIM} has ca. 297K parameters that takes ca. 10 MB. The average training time per path per epoch is 6.2 ms. In testing, the runtime is 4.2 ms per path, which is within a reasonable response time. In additon, it takes 14.2 ms for Aalborg city and 11.1 ms for Harbin city to run top-$k$ shortest path finding.

\subsubsection{How \emph{PIM} work on bigger graphs?}

\emph{PIM} takes as input a path and learns a representation for the path. Thus, since the input of \emph{PIM} is a path, not a graph, the size of the graph does not affect \emph{PIM}, but the path length (i.e., the number of vertices in a path) affects. In the experiments, we consider paths with up to 215 vertices (corresponding to ca. 40 km). 

We have also worked on multi-objective route planning. Here, the goal is to identify the ``best'' path, e.g., the Pareto-optimal path, which becomes more challenging on bigger graphs as the search space for finding the best path becomes much larger. This is different from \emph{PIM}, where paths are provided as inputs directly, e.g., already returned by routing planning algorithms.   

The curriculum negative sampling relies on the top-k diversified shortest path algorithm to generate negative paths. The top-k diversified shortest path algorithm is tested on graphs with up to 2.4 million nodes. Thus, it is able to support large graphs. 

Other graph embedding methods can work with graphs with more than one million vertices. Specifically,  \emph{node2vec} is tested on graphs with 1 million nodes and \emph{GMI} is tested on graphs with 10 million nodes. 

\section{Conclusions}
We study unsupervised path representation learning without using task-specific labels. 
We propose a novel contrastive learning framework Path InfoMax (\emph{PIM}), including a curriculum 
negative sampling strategy to generate a small number of negative paths and a training mechanism that jointly learns distinguishable path representations from both a global and a local view. 
Finally, we conduct experiments on two datasets with two downstream tasks. Experimental results show that \emph{PIM} outperforms other unsupervised methods and, as a pre-training method, \emph{PIM} is able to enhance supervised path representation learning. 

\section*{Acknowledgments}
This  work  was  supported by Independent  Research Fund Denmark under agreements 8022-00246B and 8048-00038B, the VILLUM FONDEN under agreement 34328, and the Innovation Fund Denmark centre, DIREC.

\bibliographystyle{unsrtnat}
\bibliography{references}  

\end{document}